\documentclass[conference]{IEEEtran}
\usepackage{amsmath}
\usepackage{color}
\usepackage{array}
\usepackage{stfloats}
\usepackage{amsthm} 
\usepackage{amssymb}
\usepackage{float}
\usepackage{nccmath}
\usepackage{graphicx}
\usepackage[linesnumbered,ruled]{algorithm2e}
\usepackage{setspace}
\usepackage{booktabs}
\usepackage{subcaption}
\usepackage{mwe}
\usepackage{cite}
\usepackage{amsmath,amssymb,amsfonts}
\usepackage{graphicx}
\usepackage{textcomp}
\usepackage{xcolor}
\usepackage[utf8]{inputenc} 
\usepackage{kotex} 

\usepackage{algorithmicx}
\usepackage{xspace}
\def\BibTeX{{\rm B\kern-.05em{\sc i\kern-.025em b}\kern-.08em
    T\kern-.1667em\lower.7ex\hbox{E}\kern-.125emX}}
\begin{document}
\newcommand{\BfPara}[1]{{\noindent\bf#1.}\xspace}
\newcommand\mycaption[2]{\caption{#1\newline\small#2}}
\newcommand\mycap[3]{\caption{#1\newline\small#2\newline\small#3}}





\title{Privacy-Preserving Deep Learning Computation for Geo-Distributed Medical Big-Data Platforms}

\author{\IEEEauthorblockN{Joohyung Jeon$^\dag$, Junhui Kim$^\dag$, Joongheon Kim$^\dag$, Kwangsoo Kim$^\ddag$, Aziz Mohaisen$^\ast$, 
and Jong-Kook Kim$^\S$}
\IEEEauthorblockA{$^\dag$\textit{School of Computer Science and Engineering, Chung-Ang University, Seoul, Korea} \\
$^\ddag$\textit{Biomedical Research Institute \& Center for Medical Innovation, Seoul National University Hospital, Seoul, Korea}\\
$^\ast$\textit{Department of Computer Science, University of Central Florida, Orlando, FL, USA}\\
$^\S$\textit{School of Electrical Engineering, Korea University, Seoul, Korea}
}
}

\maketitle

\begin{abstract}
This paper proposes a distributed deep learning framework for privacy-preserving medical data training. In order to avoid patients' data leakage in medical platforms, the hidden layers in the deep learning framework are separated and where the first layer is kept in platform and others layers are kept in a centralized server. Whereas keeping the original patients' data in local platforms maintain their privacy, utilizing the server for subsequent layers improves learning performance by using all data from each platform during training. 





\end{abstract}

\section{Introduction}
Artificial intelligence and deep learning computations are widely used in many areas. Among them, deep learning for medical applications is one of the most remarkable applications, where deep learning algorithms are directly utilized for medical applications, e.g., learning-based abnormality detection in medical imaging, statistical inference for public health, and deep learning based preventive medicine~\cite{mit2,tsmc15}. 

Medical images are large and sensitive by nature, making scalability and privacy two pressing issues in applying deep learning to the problem at hand. As such, developing efficient and privacy-preserving approaches for medical applications is of a paramount importance. To address those challenges, this work proposes a system-wide enhancement for \textit{patient privacy aware} computation with medical patients' big-data~\cite{icaiic19}. 


In geo-distributed medical platforms (hospitals,  biomedical institutes, etc.), there is a lot of patients' data, so-called big-data, and to conduct deep learning on such data all of it typically should be in the same space for training using supervised learning algorithms. However, gathering and transporting patient data is strictly regulated by  laws due to privacy. As a result, each medical platform conducts computations with its own local data, leading to \textit{overfitting}. Moreover, the amount of data in each platform is not equal, leading to \textit{data imbalance} and associated learning issues~\cite{icaiic19}.  To deal with those issues, this paper proposes a distributed framework which trains models with the data stored in each platform \textit{locally} for preserving privacy while maintaining learning  accuracy.


\begin{figure}[!h]
    \centering
    \includegraphics[width=0.750\columnwidth]{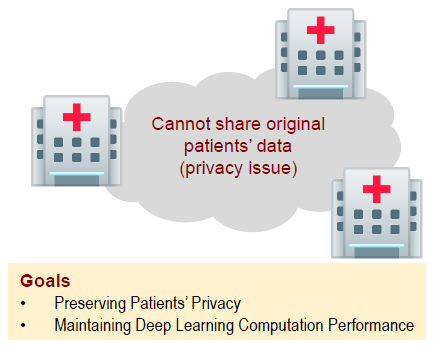}
    \caption{Deep learning computation over geo-distributed medical platforms under privacy-preserving requirements.}
    \vspace{-3mm}
    \label{fig:dsn1}
\end{figure}


\section{Proposed Learning Framework}


\BfPara{Related Work}
The \textit{de facto} standard for privacy-preserving deep learning  is~\cite{federated}, which works as follows. First of all, distributed platforms download the learning model from a centralized server and train the model by their own \textit{local} data. The updated model parameters/weights in each platform are the aggregated to the server. Each platform downloads the model from the server again and this process continues. Even though this approach helps in overcoming privacy leakage by avoiding raw and original data sharing, it requires a huge \textit{network bandwidth} in order to share models and parameters, as well as conducting \textit{computation} in each platform. Moreover, this problem is manifested further when the model becomes deeper and larger, an unavoidable fate in medical deep learning applications for extremely precise patient diagnosis. 

\begin{figure}[!t]
    \centering
    \includegraphics[width=0.90\columnwidth]{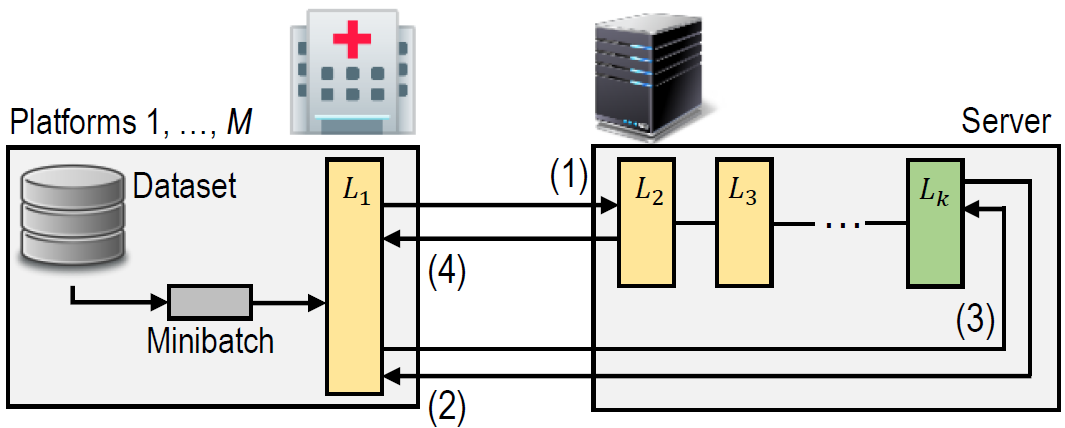}
    \caption{System model where $L_1,\cdots,L_{k-1}$ are hidden layers and $L_{k}$ is an output layer.}
    \label{fig:systemmodel}
    \vspace{-3mm}
\end{figure}

\BfPara{Proposed Distributed Deep Learning Framework}
The overarching goal of the proposed framework is to preserve privacy in deep learning while reducing network bandwidth usage and maintaining computing efficiency. 
In our framework, multiple geo-distributed medical platforms and a single central server exist, where the server is required for deep learning training with \textit{global} information from all platforms. As shown in Fig.~\ref{fig:systemmodel}, each platform has its own local dataset and trains it with a predefined minibatch (denoted by $s_{k}$ for platform $k$). 
Moreover, each platform has the first hidden layer (denoted by $L_{1}$). Since the original data is not directly shared, i.e., the data is shared in the form of the results of $L_{1}$, the server cannot look at the original data, thus patients privacy can be preserved.  
The server has the other hidden layers, i.e., $L_{2},\cdots,L_{k-1}$, and the output layer $L_{k}$. Since the server gathers the data from all platforms, the effect of training with all data can be obtained (while preserving patients' privacy).

\begin{figure}[!t]
    \centering
    \includegraphics[width=1.00\columnwidth]{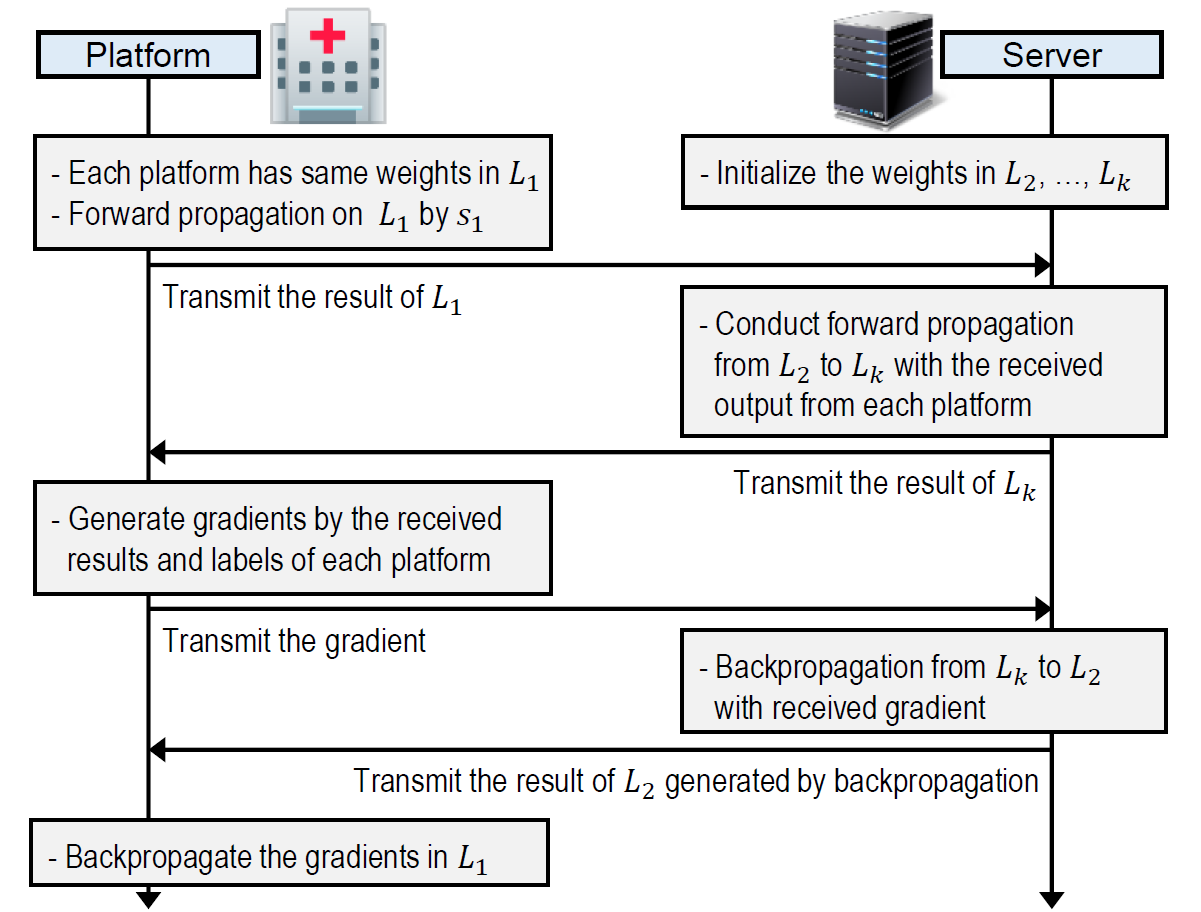}
    \caption{Flowchart for the proposed learning framework.}
    \label{fig:communication}
    \vspace{-2mm}
\end{figure}

The detailed workflow is presented in Fig.~\ref{fig:communication}. 
First, we initially postulate that each platform has the same weights in $L_{1}$ and the server initializes the weights in $L_{2},\cdots,L_{k}$.
After that, each platform conducts forward propagation on $L_{1}$ by the size of the minibatch $s_{1}$; and 
transmits the results of $L_{1}$ computation. When the server receives them, it continues forward propagation from $L_2$ to $L_k$ with the received output from each platform, i.e., the results of $L_1$, and then the results of $L_k$ are transmitted back to each platform. Based on this result, each platform generates gradients. The computed gradient values are transmitted to the server, and the server conducts backpropagation from $L_k$ to $L_2$ based on the gradient. After that, the server transmits the results of $L_2$ to each platform, and each platform finally backpropagates the gradients in $L_2$.

Finally, it can be seen that the entire procedure contains four communication processes as also illustrated in Fig.~\ref{fig:systemmodel}. 

In this distributed computation, we can additionally handle the \textit{data imbalance problem} which happens when each platform has different number of patients records. It is obvious that this introduces certain amounts of bias in training. 
In order to mitigate this data imbalance problem, the minibatch size in each platform can be adjusted as the proportion of the amount of \textit{local} data in each platform. 

\section{Primary Evaluation Results}
\begin{figure}[!t]
    \centering
    \includegraphics[width=1.05\columnwidth]{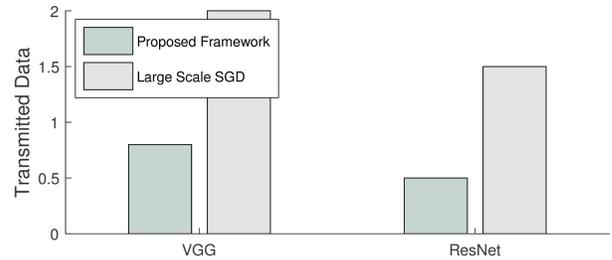}
    \caption{Communication bandwidth evaluation.}
    \label{fig:graph}
    \vspace{-3mm}
\end{figure}

In order to evaluate the communication overheads, we measure the transmitted data with our proposed method and our comparing method, i.e., \textit{large scale stochastic gradient descent (SGD)} which is one of the well-known methods~\cite{SGD}. Fig.~\ref{fig:graph} shows the amounts of the transmitted data while training VGG and ResNet with CIFAR-10 and CIFAR-100. While training the VGG model, $0.8$\,GB data are communicated in the proposed framework with $95$\,\% accuracy whereas the Large Scale SGD is with $2$\,GB data transmission and $55$\,\% accuracy. In addition, training ResNet shows a pronounced communication cost reduction. With our proposed framework, $0.5$\,GB data are transmitted with $75$\,\% accuracy whereas the Large Scale SGD transmits $1.5$\,GB with $10$\,\% accuracy.

\section{Summary and Future Work}
In this paper, we propose a distributed deep learning framework for privacy-preserving computation. Based on the given deep neural network, the hidden layers are separated and then the first layer is left in each platform where the other layers are in a centralized server. By doing this, the original/raw patients' data in each medical platform is not leaked during training, thus ensuring privacy. Furthermore, utilizing the centralized server helps to improve learning performance by using all data from individual platforms during training. As future work, implementing this framework in geo-distributed hospitals (i.e., Seoul National University Hospitals) is anticipated. 

\section*{Acknowledgment}
This work was supported by IITP (2017-0-00068 and 2018-0-00170) and also by National Research Foundation of Korea (2016R1C1B1015406 and 2017R1A4A1015675). 
J. Kim is a corresponding author (e-mail: joongheon@gmail.com).

\end{document}